\documentclass[letterpaper, 10 pt, journal, twoside]{ieeetran} 
\usepackage{graphicx}
\usepackage{amsmath}
\usepackage{bbding}
\usepackage{pifont}
\usepackage{wasysym}
\usepackage{amssymb}
\usepackage{hyperref}
\usepackage{bbm}
\usepackage{threeparttable} 
\usepackage{cite} 
\usepackage{booktabs}
\usepackage{xcolor}
\usepackage{multirow}
\usepackage{blindtext}
\usepackage{bm}
\usepackage{subcaption}
\hypersetup{hidelinks}
\usepackage{amsmath,amssymb,mathtools}
\usepackage[linesnumbered, ruled]{algorithm2e}

\SetKwInput{KwInput}{Input}
\SetKwInput{KwOutput}{Output}

\usepackage{cite}

\IEEEoverridecommandlockouts                              
\usepackage{geometry}
\usepackage{afterpage}
\usepackage{mwe}
\title{RNGDet++: Road Network Graph Detection by Transformer with Instance Segmentation and Multi-scale Features Enhancement}
\author{Zhenhua Xu, \IEEEmembership{Graduate Student Member, IEEE}, 
Yuxuan Liu, \IEEEmembership{Graduate Student Member, IEEE},\\  Yuxiang Sun, \IEEEmembership{Member, IEEE}, Ming Liu, \IEEEmembership{Senior Member, IEEE}, and Lujia Wang,  \IEEEmembership{Member, IEEE} %

\thanks{Manuscript received September 21, 2022; Revised December 23, 2022; Accepted March 1, 2023. This paper was recommended for publication by Editor Pauline Pounds upon evaluation of the Associate Editor and Reviewers' comments. This work was supported in part by Guangdong Basic and Applied Basic Research Foundation under Grants 2021B1515120032 and 2022A1515010116, in part by Foshan-HKUST Project under Grant FSUST20-SHCIRI06C, and in part by Hetao Shenzhen-Hong Kong Science and Technology Innovation Cooperation Zone Project under Grant HZQB-KCZYB-2020083.
\textit{(Corresponding author: Lujia Wang.})}
\thanks{Zhenhua Xu and Yuxuan Liu are with The Hong Kong University of Science and Technology (email: zxubg@connect.ust.hk, yliuhb@connect.ust.hk).}
\thanks{Yuxiang Sun is with the Department of Mechanical Engineering, The Hong Kong Polytechnic University, Hung Hom, Kowloon, Hong Kong (e-mail: yx.sun@polyu.edu.hk, sun.yuxiang@outlook.com).}
\thanks{Ming Liu is with The Hong Kong University of Science and Technology (Guangzhou), Nansha, Guangzhou, 511400, Guangdong, China, and also with The Hong Kong University of Science and Technology, Hong Kong SAR, China, and also with HKUST Shenzhen-Hong Kong Collaborative Innovation Research Institute, Futian, Shenzhen. (email: eelium@ust.hk).}
\thanks{Lujia Wang is with The Hong Kong University of Science and Technology, and also with Clear Water Bay Institute of Autonomous Driving (Shenzhen) (email: eewang@ust.hk).} 
\thanks{Digital Object Identifier (DOI): see top of this page.} 
} 

\IEEEaftertitletext{\vspace{-1.0\baselineskip}}
\begin{document}
\bstctlcite{IEEEexample:BSTcontrol}
\maketitle
\begin{abstract}      
The road network graph is a critical component for downstream tasks in autonomous driving, such as global route planning and navigation. In the past years, road network graphs are usually annotated by human experts manually, which is time-consuming and labor-intensive. To annotate road network graphs effectively and efficiently, automatic algorithms for road network graph detection are demanded. Most existing methods either adopt a post-processing step on semantic segmentation maps to produce road network graphs, or propose graph-based algorithms to directly predict the graphs. However, these works suffer from hard-coded algorithms and inferior performance. To enhance the previous state-of-the-art (SOTA) method RNGDet, we add an instance segmentation head to better supervise the training, and enable the network to leverage multi-scale features of the backbone. Since the new proposed approach is improved from RNGDet, we name it RNGDet++. Experimental results show that our RNGDet++ outperforms baseline methods in terms of almost all evaluation metrics on two large-scale public datasets. Our code and supplementary materials are available at \url{https://tonyxuqaq.github.io/projects/RNGDetPlusPlus/}.


\vspace{0.25cm}

\end{abstract}

\section{Introduction}
\IEEEPARstart{T}{he} vector map of road elements, including high-definition (HD) maps and standard-definition (SD) maps, is important for autonomous vehicles. The road network graph is a kind of SD map that records reliable position and topology information of drivable roads. It is a fundamental component for downstream tasks of autonomous driving, such as global route planning and navigation \cite{liu2021role,christensen2021autonomous}. Autonomous vehicles can query prior road network graphs to find an optimal road path to reach a destination, especially in complicated urban scenarios. In addition, road network graphs can be applied in navigation tasks, such as using Google maps to navigate ourselves in daily life. Usually, a road network graph consists of vertices and edges, where vertices represent key points of the road network (e.g., road ends and road intersections) and edges represent road segments. Since road network graphs can cover a large area, such as a city or even a country, manually annotating it is time-consuming and labor-intensive, which severely increases the cost and hinders the wide applications of autonomous vehicles. Therefore, methods that can automatically detect the road network graph are of great interest to the research community.

Since road network graphs only contain road-level information, the detection for them does not require images with very high resolution. So, aerial images obtained by unmanned aerial vehicles (UAVs) or satellites \cite{mnih2010learning} are sufficient for this task. They are much cheaper and easier to access than the images/point clouds collected by vehicle-mounted sensors. In this paper, the road network graph is detected from large aerial images at the resolution of 1m/pixel. There are many existing works in this area, which could be classified into three categories. 
The first category is based on semantic segmentation \cite{hu2014road,shi2013spectral,unsalan2012road,cheng2016road,batra2019improved,mattyus2017deeproadmapper,mnih2010learning,etten2020city,bandara2022spin,cheng2017automatic,zhou2021split}. These works first predict the semantic segmentation mask of road networks and then extract the graphs by post-processing. 
The second category of works has two stages and can directly obtain the road network graph without complicated post-processing \cite{xu2022csboundary,he2020sat2graph,bahl2022single}. These works first calculate graph vertices by predicting the vertex heatmap, and then predict graph edges by connecting obtained graph vertices. 
The last category treats the graph detection task as a Markov decision process (MDP) problem and proposes an iterative decision-making algorithm to detect graphs \cite{bastani2018roadtracer,tan2020vecroad,xu2022rngdet,li2018polymapper}. Starting from predicted initial candidates, these works train an agent network that can detect the road network graph by iterations, which behaves in a similar way to human experts. Among them, Road Network Graph Detector (RNGDet) \cite{xu2022rngdet}  proposes and trains a DETR-like (Detection by Transformer) \cite{carion2020end} transformer network to track and detect the road network graph, which presents the state-of-the-art (SOTA) performance so far. However, RNGDet does not fully make use of multi-scale features extracted by the CNN backbone, which restricts the further improvement of this network. 

In this letter, we propose RNGDet++, a novel approach that directly detects road networks in the graph format. Compared with RNGDet, RNGDet++ can better use the multi-scale features of the backbone and presents superior results in our task. Besides, we add an instance segmentation head into the network to better supervise the training, which enhances our robustness and performance. RNGDet++ is trained by imitation learning. We conduct comparative experiments and ablation studies on the city-scale dataset released by Sat2Graph \cite{he2020sat2graph}, and the SpaceNet dataset \cite{van2018spacenet}. Our contributions are listed as follows:
\begin{itemize}
    \item We propose RNGDet++, a novel approach that can make full use of the multi-scale features to effectively detect road networks in the graph format.
    \item We add an instance segmentation head to the network, which can better supervise the training of the network and improve the performance.
    \item We evaluate our RNGDet++ and all the baseline methods on two large-scale public datasets. Our RNGDet++ presents superior results.
    \item We open-source our code and release the data at \url{https://tonyxuqaq.github.io/projects/RNGDetPlusPlus/}.
\end{itemize}

\section{Related Works}
\subsection{Graph Detection of Simple Road Elements from Bird's-Eye View}
Simple road elements refer to those with simpler topology, such as road boundaries \cite{liang2019convolutional,xu2021topo,xu2022csboundary}, road curbs \cite{zhxu2021icurb,xu2021cp} and road lanelines \cite{homayounfar2018hierarchical,homayounfar2019dagmapper,li2022hdmapnet}. Under common circumstances, these road elements do not have complicated merges, splits, or intersections, thus the detection of the graph of these road elements is less difficult. The input data of past works is bird's-eye view (BEV) images which are either aerial images \cite{zhxu2021icurb,xu2021topo,xu2022csboundary} or BEV images of the pre-built map \cite{liang2019convolutional,homayounfar2018hierarchical,homayounfar2019dagmapper,xu2022centerlinedet}. Most previous works utilize a decision-making network to iteratively detect the graph of target objects. Liang \textit{et al.} \cite{liang2019convolutional} proposed a CNN-based decision network to detect the road boundary in BEV images obtained from the pre-built point cloud map. Li \textit{et al.} \cite{homayounfar2019dagmapper} further modified the algorithm to handle simple topology changes of lanelines (e.g., split and merge) on highways. Although the aforementioned approaches could achieve satisfactory results in their specific detection tasks, they cannot be adapted to the detection task of the road network graph, since road network tends to have more complicated topology structure, such as road intersections and road overlapping (e.g., overpasses). Therefore, more powerful and robust algorithms are demanded.

\subsection{Road Network Detection from Aerial Images}
With the fast development of aerial imaging techniques, high-resolution aerial images from all over the globe could be easily accessed nowadays. Thus, most past works on road network detection take aerial images as input \cite{hu2014road,shi2013spectral,unsalan2012road,cheng2016road,batra2019improved,mattyus2017deeproadmapper,mnih2010learning,etten2020city,bandara2022spin,cheng2017automatic,zhou2021split,bastani2018roadtracer,tan2020vecroad,li2018polymapper,xu2022rngdet,bahl2022single}. They could be classified into three categories: (1) Segmentation-based approaches \cite{hu2014road,shi2013spectral,unsalan2012road,cheng2016road,batra2019improved,mattyus2017deeproadmapper,mnih2010learning,etten2020city,bandara2022spin,cheng2017automatic,zhou2021split}. This category of approaches first predict the semantic segmentation map of road networks, and then conduct post-processing algorithms (e.g., skeletonization and binarization) to extract the graph. However, they usually have inferior topology correctness, especially when road intersections or road overlappings are encountered; (2) Two-stage-graph-based approaches \cite{he2020sat2graph,bahl2022single}. He \textit{et al.} \cite{he2020sat2graph} proposed a two-stage algorithm Sat2Graph to directly predict the graph of road networks without complicated hard-code post-processing. The authors first predicted the heatmap of road network graph vertices and extracted vertex coordinates by processing algorithms. Then, based on predicted graph vertices, they designed an encoding scheme to demonstrate graph edges by projecting the input aerial image into an 18-D tensor. A deep neural network is trained to predict the 18-D encoding tensor of the input image, and the graph edges could be calculated by decoding the predicted encoding tensor. Sat2Graph presents quite promising results, but it is not end-to-end trainable, which degrades its final performance. Moreover, the isomorphic encoding issue \cite{he2020sat2graph} also restricts Sat2Graph from having better evaluation scores; (3) Iterative-graph-based approaches \cite{bastani2018roadtracer,tan2020vecroad,li2018polymapper,xu2022rngdet}. These approaches convert the road network detection task to an MDP problem, in which an agent is trained to detect the road network graph vertex by vertex iteratively. It is believed that RoadTracer proposed by Bastani \textit{et al.} \cite{bastani2018roadtracer} is the first work belonging to this category of approaches. RoadTracer trains a CNN-based decision network to control an agent to explore the road network by iterations. At each step, the network predicts the moving direction of the next step, and the agent moves in the predicted direction by a fixed distance. Inspired by RoadTracer, Xu \textit{et al.} \cite{xu2022rngdet} proposed a DETR-like network RNGDet to detect the road network graph. RNGDet achieves the SOTA performance. At each step, RNGDet directly predicts the coordinates of vertices in the next step, so that RNGDet can have an adjustable step length and handle road intersections with arbitrary numbers of incident roads. However, RNGDet only utilizes the feature of the deepest layer of the backbone, leaving multi-scale features not fully used, which prevents further improvement.

\begin{figure*}[t]
\centering
    \includegraphics[width=\linewidth]{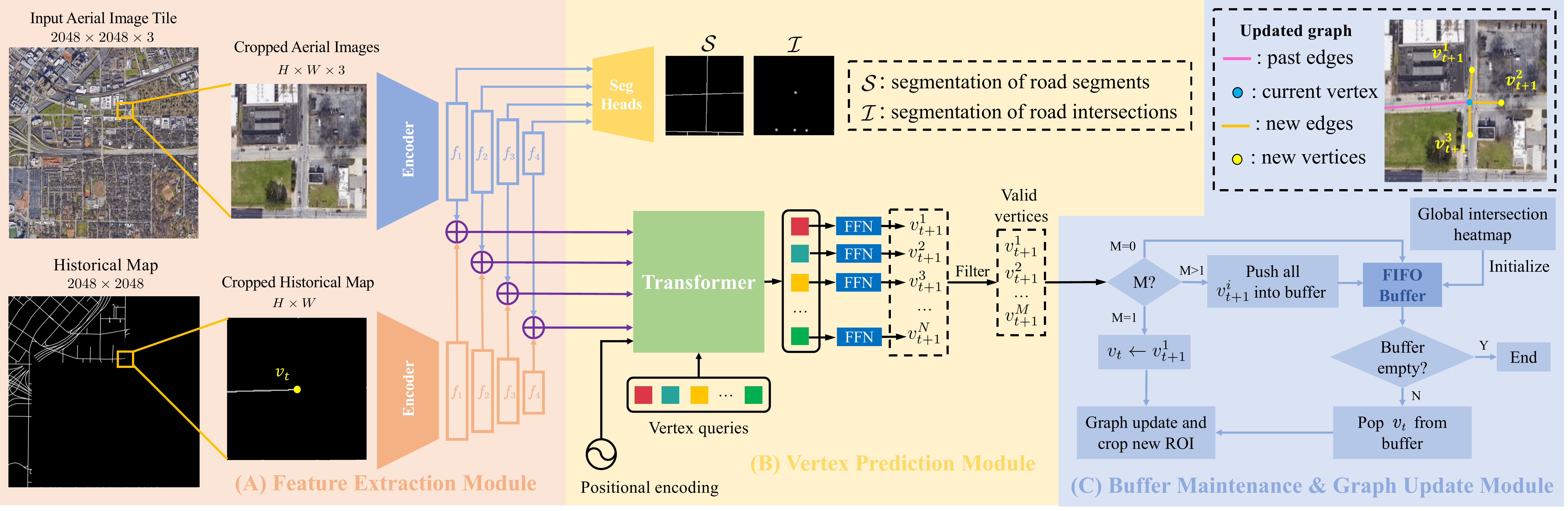}
  \caption{System diagram of RNGDet++. In this diagram, RNGDet++ conducts single-step processing at time $t$. RNGDet++ mainly consists of three modules: (A) Feature extraction module. With the RGB aerial image and the binary historical map as input, this module first crops ROIs centering at $v_t$, and then extracts multi-scale deep features of ROIs by two backbones. (B) Vertex prediction module. Based on extracted multi-scale features, this module predicts two semantic segmentation maps $\mathcal{S}$, $\mathcal{I}$, vertices in the next step $\{v_{t+1}^i\}_{i=1}^M$, as well as instance segmentation maps of all valid vertices $\{(\mathcal{S_I})_{t+1}^i\}_{i=1}^M$. (C) Buffer maintenance \& graph update module. After obtaining vertices in the next step, RNGDet++ updates the graph $G$, and controls the agent to make corresponding actions. The agent keeps repeating the above steps until the buffer is empty. When the buffer is empty, RNGDet++ stops and outputs the predicted road network graph $G$. This figure is best viewed in color. Please zoom in for details.}
  \label{system diagram}
\end{figure*}

\subsection{Detection by Transformer}
Compared with CNN, transformer \cite{vaswani2017attention} can better handle variant-length input, and capture global relationships between patches of the input image. Transformer-based detection framework Detection by Transformer (DETR) is first proposed by Carion \textit{et al.} \cite{carion2020end}. Compared with previous detection works, DETR is more simple, effective, and end-to-end trainable. Taken as input images, DETR directly outputs a fixed-length vector encoding certain information about each candidate object. By modifying the definition of the output vector as well as the transformer network, DETR is adopted to handle various different detection tasks, such as line segment detection \cite{xu2021line}, road centerline detection \cite{can2021structured}, and road network graph detection \cite{xu2022rngdet}. Even if most DETR-like approaches present satisfactory results for a specific task, they only utilize a single feature layer obtained by the backbone, while the multi-scale features are not fully used. In this paper, RNGDet++ is proposed to conquer this problem by using multi-scale features for both training and inference to further enhance the final performance.

\section{Methodology}

\subsection{Overview}
In this letter, we propose RNGDet++ to detect the graph of road networks for downstream autonomous driving applications. Compared with RNGDet, RNGDet++ makes full use of multi-scale backbone features and adds an instance segmentation head to better supervise the training phase. Suppose the road network graph is $G=(V,E)$, where $V$ is a set of key points of the road network as vertices and $E$ is a set of road segments as edges. Taken as input large aerial image tiles, the task of RNGDet++ is to predict the road network graph $G$. The system diagram of RNGDet++ is displayed in Fig. \ref{system diagram}.

RNGDet++ controls an agent to iteratively detect the road network graph, whose current coordinates are denoted by $v_t$, where $t$ is the current time stamp. Since the input aerial image tile (i.e., $I$) is usually very large, such as $2048\times2048$ or $4096\times4096$, considering limited computation resources, RNGDet++ processes a $128\times128$ region of interest (ROI) at one time. To provide the agent with historical information, the rasterized graph detected by RNGDet++ so far is recorded as the historical map $H$. $H$ is represented by a binary image whose size is the same as that of $I$. $H$ is obtained by rasterizing the vector format historical graph into an image so that it can be processed by the CNN backbone together with $I$. Centering at $v_t$, the image ROI (i.e., $I_R$) and historical map ROI (i.e., $H_R$) are cropped on $I$ and $H$, respectively. Taken as input $I_R$ and $H_R$, RNGDet++ extracts the multi-scale deep features by two ResNet \cite{he2016deep} backbones. The $i$-th feature layer is denoted by $f_i$, and larger $i$ indicates a deeper feature layer. 

With a feature pyramid network (FPN) \cite{lin2017feature}, RNGDet++ predicts the segmentation of road segments (i.e., $\mathcal{S}$) and the segmentation of road intersection points (i.e., $\mathcal{I}$). The DETR-like transformer network makes full use of multi-scale backbone features, and predicts coordinates (i.e., $\{v_{t+1}^i\}_{i=1}^N$) and valid probability (i.e., $\{p_{t+1}^i\}_{i=1}^N$) of $N$ vertices in the next step. These predicted vertices are then filtered by removing those with low valid probability $p_{t+1}^i$ and RNGDet++ finally obtains $M$ valid vertices in the next step.

RNGDet++ maintains a first-in-first-out (FIFO) buffer saving initial candidates, which are initial vertices to initialize the iteration of the agent. These initial candidates may come from local peaks of the global segmentation heatmap of road intersections, or from breakpoints of the agent iteration. Based on the number of valid vertices $M$ in the next step, the agent takes different actions to update the graph. If $M=1$, there is only one vertex $v_{t+1}^1$ in the next step, and the agent directly moves to $v_{t+1}^1$; if $M=0$, the agent pops a new initial candidate from the buffer; if $M>1$, it indicates that road intersections are met, so the agent pushes all $\{v_{t+1}^i\}_{i=1}^M$ into the buffer and pops a new initial candidate from it. After this, the agent crops new ROIs, predicts new vertices, and repeats the aforementioned steps until the buffer is empty.

\begin{figure*}[t]
\centering
    \includegraphics[width=\linewidth]{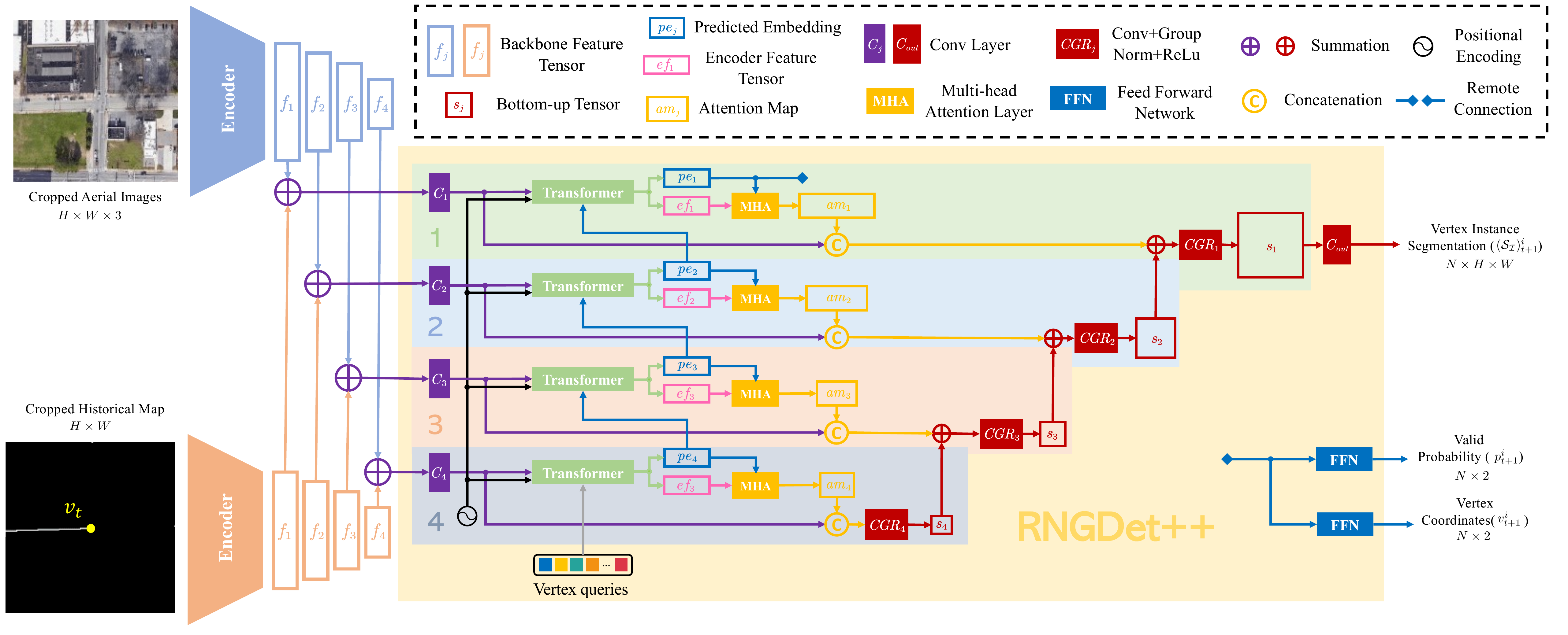}
  \caption{Network structure of RNGDet++. RNGDet++ can make use of all four levels of features while RNGDet only utilizes the 4-th feature layer. At each layer, the transformer predicts an embedding tensor $pe_i$ (blue box) of each input vertex query, and outputs the feature tensor extracted by the transformer encoder $ef_i$ (pink box). $pe_1$ is fed to feed-forward networks (FFNs) to predict the coordinates and valid probability. A multi-head attention layer (orange rectangle) is used to predict the attention map $am_i$ (orange box) based on $pe_i$ and $ef_i$. The predicted attention map is then sent to an FPN (red elements) to predict the instance segmentation map. This figure is best viewed in color. Please zoom in for details. }
  \label{multi-scale}
\end{figure*}

\subsection{Feature Extraction}
Centering at the current coordinate of the agent (i.e., $v_t$), RNGDet++ crops $I_R$ and $H_R$ on the input aerial image and the historical map. $I_R$ and $H_R$ provide the agent with visual information and historical information, respectively. RNGDet++ extracts multi-scale deep features of these two ROIs by two ResNet backbones. Each ResNet backbone can obtain four layers of features denoted as $f_i,i\in(1,2,3,4)$. A larger $i$ indicates a deeper feature layer. The extracted features obtained by two backbones are added for feature fusion. 
 
\subsection{Multi-scale Feature Fusion}
The main difference between RNGDet++ and the previous RNGDet is that RNGDet++ can make use of the multi-scale backbone features while RNGDet only utilizes a single feature layer. Suppose the ResNet backbone obtains four levels of features denoted by $\{f_i\}_{i=1}^4$, and larger $i$ indicates a deeper level feature tensor. RNGDet only uses the deepest layer $f_4$ for vertex prediction while tensors at other levels are ignored, which causes information loss. Inspired by UNet \cite{ronneberger2015u} and FPN \cite{lin2017feature}, the proposed RNGDet++ makes predictions based on features extracted at all four levels, which can better capture the deep feature of the input images to predict the vertex in the next step. Please refer to our supplementary document for a detailed comparison between RNGDet and RNGDet++.

Transformers are trained to process multi-scale features. Taken as input $f_i$ and a vertex query, a shared transformer predicts an embedding tensor $pe_i$. $pe_1$ is sent to a feed-forward network (FFN) for the prediction of one vertex in the next step. Together with the output of transformer encoder $ef_i$, $pe_i$ is used to calculate the attention map $am_i$. Both $\{am_i\}_{i=1}^4$ and $\{f_i\}_{i=1}^4$ are fed to an FPN segmentation head for instance segmentation of road segments ahead. The network structure of RNGDet++ is shown in Fig. \ref{multi-scale}.

\subsection{Vertex Prediction}
With the extracted multi-scale deep features as input, RNGDet++ outputs several predictions by different heads.
 
\subsubsection{Semantic Segmentation}
RNGDet++ predicts the semantic segmentation maps of road segments (i.e., $\mathcal{S}$) and road intersection points (i.e., $\mathcal{I}$) with an FPN segmentation head. $\mathcal{S}$ helps the network to learn the feature of the road network, while $\mathcal{I}$ enables the network to be better aware of road intersections, which improves the performance of RNGDet++ to detect road networks with complicated intersections. The semantic segmentation head only takes deep features of the input aerial image as input and ignores that of the historical map.
 
\begin{figure}[!t]
  \centering
  \includegraphics[width=\linewidth]{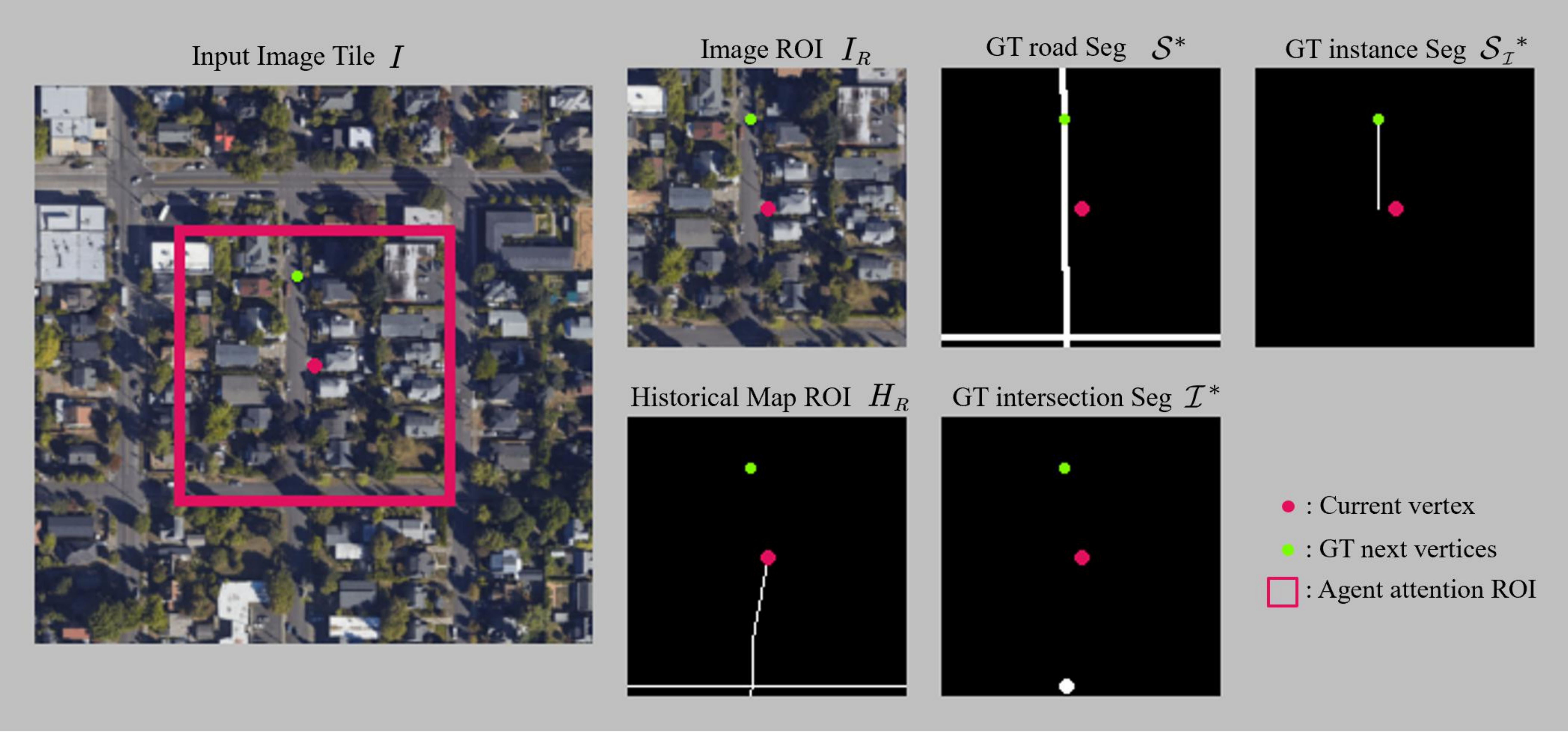}
  \caption{Visualization of the ground-truth instance segmentation mask label (top right mask). When the agent $v_t$ (pink point) is away from the right track, the instance segmentation head still supervises the agent to capture the correct road information, which improves the final performance.}
  \label{instance_seg}
\end{figure}

\begin{figure}[t]
\centering
    \includegraphics[width=\linewidth]{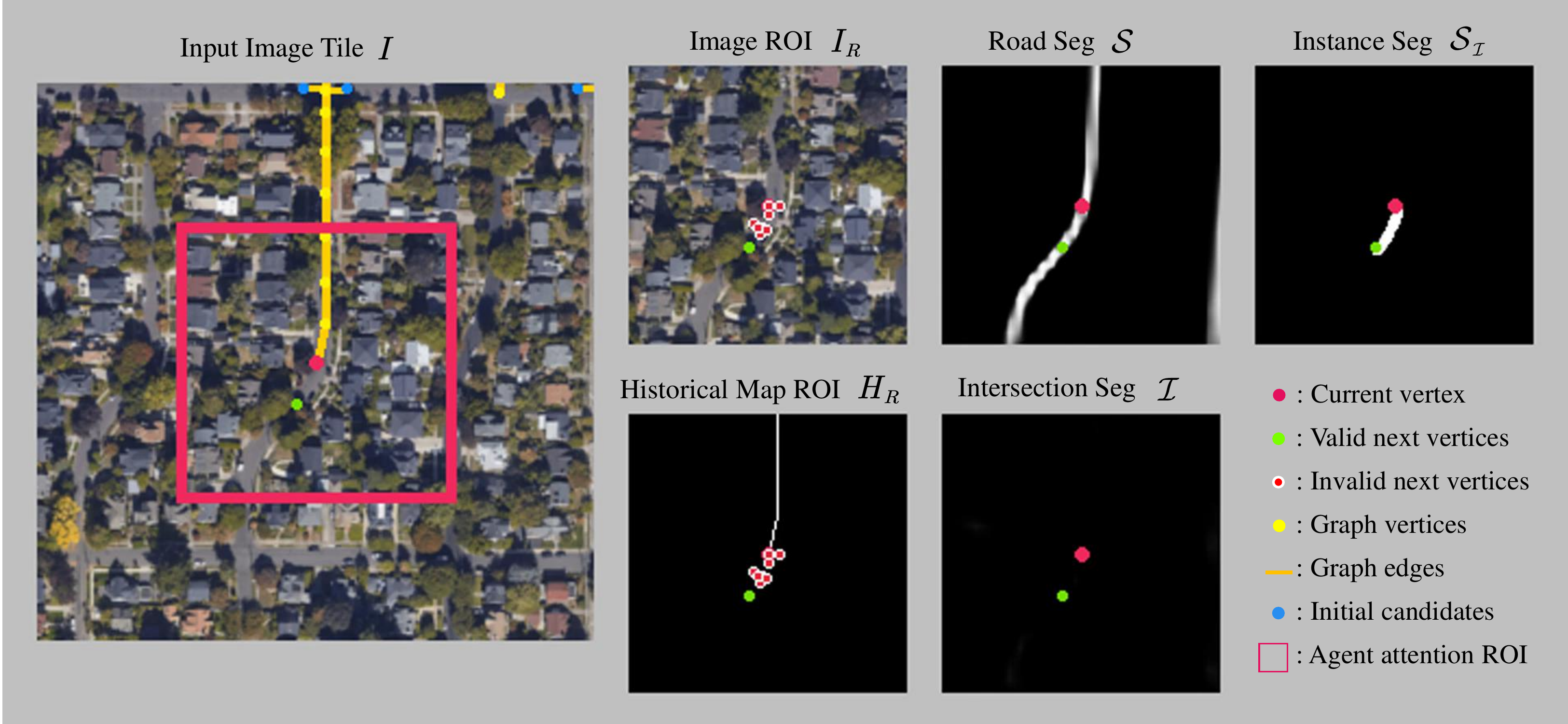}
  \caption{Visualization demonstrations for the output of RNGDet++. With ROIs $I_R$ and $H_R$ as input, RNGDet++ predicts two semantic segmentation maps $\mathcal{S}$ and $\mathcal{I}$, as well as $N$ vertices in the next step. The prediction of each vertex contains its coordinates, valid probability, and instance segmentation of the road ahead. Vertex with high valid probability is classified as valid vertex (green node), otherwise it is filtered out as invalid (red node with white margin). This figure is best viewed in color. Please zoom in for details.}
  \label{onestep}
\end{figure}

\subsubsection{Next Step Vertices}
Besides multi-scale deep features, the DETR-like transformer also takes $N$ vertex queries as input. The vertex query is a fixed-length trainable tensor, which could be treated as a slot that the transformer can utilize to make predictions. For each vertex query, the transformer predicts the x-y coordinate of the vertex in the next step $v_{t+1}^i$, a valid probability $p_{t+1}^i$, and an instance segmentation map of the ahead road $(\mathcal{S_I})_{t+1}^i$. If $p_{t+1}^i$ is larger than a threshold, the predicted vertex is classified as valid and will be used to update the graph. Suppose $v^*$ is the vertex projected from a vertex $v$ onto the ground-truth road network. $(\mathcal{S_I})_{t+1}^i$ is the road segment ahead of $v_t$ that connects $v_t^*$ and $(v_{t+1}^i)^*$, which could better supervise the training phase of the network and improve the reasoning ability of the agent. An example is visualized in Fig. \ref{instance_seg} to show how we define instance segmentation labels.
 
Since there are $N$ input vertex queries, the transformer predicts $N$ vertices in the next step. After filtering those vertices with low valid probability, we finally have $M$ valid vertices in the next step. The visualization of RNGDet++ outputs is shown in Fig. \ref{onestep}.
 
\subsection{Buffer Maintenance \& graph update}
During the iteration of RNGDet++, we maintain a FIFO buffer to save initial candidates. An initial candidate is one initial vertex that the agent starts the iteration. At the very beginning, we predict the global segmentation heatmap of road intersections by merging $\mathcal{I}$ of ROIs that cover the whole $I$, and push all local peak points of the segmentation map into the buffer as initial candidates. Then, we pop one initial candidate from the buffer, crop ROIs centering at the newly popped initial candidate and predict vertices in the next step. Based on the number of valid predicted vertices $M$, we take different actions to maintain the buffer and update the graph:
\begin{itemize}
    \item $M=0$. No road is ahead. Thus, the agent stops processing the current road instance, and pops a new initial candidate from the buffer to work on.
    \item $M=1$. A single road is ahead. The agent directly moves to $v_{t+1}^i (i=1)$ and continues the iteration without operating on the buffer.
    \item $M>1$. Multiple roads are met (e.g., road split or road intersections). The agent pushes all $v_{t+1}^i$ into the buffer as new initial candidates, and pops a new vertex from the buffer to work on.
\end{itemize} 
RNGDet++ keeps repeating the aforementioned steps until the buffer is empty. If the buffer is empty, RNGDet++ completes the detection task of the current input aerial image and outputs the predicted $G$. The working pipeline of RNGDet++ is visualized in module C of Fig. \ref{system diagram}.

\subsection{Training Data Sampling}
The training of RNGDet++ relies on imitation learning. We create an expert by using the breadth-first search (BFS) graph traversal algorithm proposed in \cite{xu2022rngdet}. The expert can annotate the road network graph vertex by vertex in the same way as human experts. For better robustness, we add even-distributed noise to expert trajectories. 

Then the task of RNGDet++ is to mimic the behavior of the expert and try to learn its policy. In our experiment, considering the training efficiency, we use the behavior-cloning imitation learning algorithm \cite{osa2018algorithmic} to collect training samples and train RNGDet++.

\subsection{Loss Functions}
For semantic segmentation maps $\mathcal{S}$ and $\mathcal{I}$, binary cross entropy loss $\mathcal{L}_{seg}$ is utilized. To alleviate the imbalance of foreground pixels and background pixels, a larger training weight is applied for foreground pixels.

Similar to DETR \cite{carion2020end}, for vertices in the next step, RNGDet++ first matches predictions with the ground truth by conducting the Hungarian algorithm. After matching, the vertex coordinate is trained by L1 loss (i.e., $\mathcal{L}_{coord}$), and valid probability is trained by cross-entropy loss (i.e., $\mathcal{L}_{prob}$). 

For each valid predicted vertex, we calculate the binary cross entropy segmentation loss (i.e., $\mathcal{L}_{ins}$) to train the instance segmentation head.

Finally, we have the training loss $\mathcal{L}=\mathcal{L}_{seg}+\alpha\mathcal{L}_{coord}+\beta\mathcal{L}_{prob}+\gamma\mathcal{L}_{ins}$.

\section{Experimental Results and Discussions}
\subsection{Dataset}
In this paper, all experiments are conducted on the city-scale dataset released in \cite{he2020sat2graph} and the SpaceNet dataset \cite{van2018spacenet}. The city-scale dataset contains 180 RGB aerial images at $2048\times2048$ (pixels) captured from different cities across the world. This dataset provides the ground truth for road networks in the format of vector graphs. Following Sat2Graph \cite{he2020sat2graph}, we split the dataset into train/valid/test with 144/9/27 tiles, respectively. The SpaceNet dataset has 2551 RGB aerial images at $400\times400$ (pixels). Compared with the city-scale dataset, the SpaceNet dataset focuses on smaller regions. The dataset is split into train/valid/test with 2042/127/382 images, respectively. Images of both datasets have a 1m/pixel resolution. These two datasets are large and have aerial images collected from various scenarios, making them sufficient for training road network graph detection approaches and conducting evaluations comprehensively.

\begin{figure*}[t]
    \newcommand{\picvisi}[1]{8_8}
    \newcommand{\picvisii}[1]{49_5}
    \newcommand{\picvisiii}[1]{68_7}
    \newcommand{\picvisiv}[1]{99_2}
 \centering
    \begin{subfigure}[t]{0.245\textwidth}
        \begin{subfigure}[t]{\textwidth}
            \includegraphics[width=\textwidth]{img/\picvisii{_}_gt.pdf}
        \end{subfigure}\vspace{.6ex}
        \begin{subfigure}[t]{\textwidth}
            \includegraphics[width=\textwidth]{img/\picvisiii{_}_gt.pdf}
        \end{subfigure}\vspace{.6ex}
        \begin{subfigure}[t]{\textwidth}
            \includegraphics[width=\textwidth]{img/\picvisiv{_}_gt.pdf}
        \end{subfigure}\vspace{.6ex}
        \caption{Ground truth}
        \label{fig_qualitative_1st}
    \end{subfigure}
    \begin{subfigure}[t]{0.245\textwidth}
        \begin{subfigure}[t]{\textwidth}
            \includegraphics[width=\textwidth]{img/\picvisii{_}_sat.pdf}
        \end{subfigure}\vspace{.6ex}
        \begin{subfigure}[t]{\textwidth}
            \includegraphics[width=\textwidth]{img/\picvisiii{_}_sat.pdf}
        \end{subfigure}\vspace{.6ex}
        \begin{subfigure}[t]{\textwidth}
            \includegraphics[width=\textwidth]{img/\picvisiv{_}_sat.pdf}
        \end{subfigure}\vspace{.6ex}
        \caption{Sat2Graph \cite{he2020sat2graph}}
        \label{fig_qualitative_1st}
    \end{subfigure}
    \begin{subfigure}[t]{0.245\textwidth}
        \begin{subfigure}[t]{\textwidth}
            \includegraphics[width=\textwidth]{img/\picvisii{_}_RNGDet.pdf}
        \end{subfigure}\vspace{.6ex}
        \begin{subfigure}[t]{\textwidth}
            \includegraphics[width=\textwidth]{img/\picvisiii{_}_RNGDet.pdf}
        \end{subfigure}\vspace{.6ex}
        \begin{subfigure}[t]{\textwidth}
            \includegraphics[width=\textwidth]{img/\picvisiv{_}_RNGDet.pdf}
        \end{subfigure}\vspace{.6ex}
        \caption{RNGDet \cite{xu2022rngdet}}
        \label{fig_qualitative_1st}
    \end{subfigure}
    \begin{subfigure}[t]{0.245\textwidth}
        \begin{subfigure}[t]{\textwidth}
            \includegraphics[width=\textwidth]{img/\picvisii{_}_RNGDet++.pdf}
        \end{subfigure}\vspace{.6ex}
        \begin{subfigure}[t]{\textwidth}
            \includegraphics[width=\textwidth]{img/\picvisiii{_}_RNGDet++.pdf}
        \end{subfigure}\vspace{.6ex}
        \begin{subfigure}[t]{\textwidth}
            \includegraphics[width=\textwidth]{img/\picvisiv{_}_RNGDet++.pdf}
        \end{subfigure}\vspace{.6ex}
        \caption{RNGDet++}
        \label{fig_qualitative_1st}
    \end{subfigure}    
    \caption{Qualitative visualization on the city-scale dataset. (a) Ground truth road networks (Cyan lines). (b) Road network graph detected by Sat2Graph. (c) Road network graph detected by RNGDet. (d) Road network graph detected by RNGDet++. For (b)-(d), yellow points represent graph vertices and orange lines represent graph edges. For the visualization, it is found that RNGDet++ can output road network graphs with more accurate structure and correctness compared with previous works. This figure is best viewed in color. Please zoom in for details. }
    \label{vis}
\end{figure*}

\subsection{Baselines and Evaluation Metrics}
We compare our RNGDet++ with previous SOTA approaches, including four segmentation-based approaches and three graph-based approaches.

\begin{itemize}
    \item Segmentation-based baselines \cite{ronneberger2015u,mattyus2017deeproadmapper,batra2019improved,he2020sat2graph}. Unet \cite{ronneberger2015u} is a widely used classic semantic segmentation network, and we adopt it to our task.  Deep Road Mapper (DRM) \cite{mattyus2017deeproadmapper} and ImprovedRoad \cite{batra2019improved} conduct post-processing steps to refine the road network segmentation results, which achieves better performance. Deep Layer Aggregation (DLA) \cite{yu2018deep} has a more powerful backbone, which is also utilized by Sat2Graph \cite{he2020sat2graph} for feature extraction.    
    \item Graph-base baselines \cite{bastani2018roadtracer,he2020sat2graph,xu2022rngdet}. RoadTracer \cite{bastani2018roadtracer} is believed to be the first graph-based approach for the road network detection task. Sat2Graph and RNGDet are two SOTA approaches that can directly output the graph of road networks.
\end{itemize}
 
All the approaches are evaluated with the metrics used in \cite{he2020sat2graph}, including TOPO \cite{biagioni2012inferring} and average path length similarity (APLS) \cite{van2018spacenet}. Within the input aerial image, TOPO first randomly samples seed vertices on the ground truth graph and the predicted one, then compares the similarity of sub-graphs that seed vertices can reach. This metric uses precision, recall, and F1 to measure the average sub-graph similarity. APLS randomly samples a vertex pair $(v_1,v_2)$ on the ground truth graph and projects them to the predicted graph as $(\hat{v}_1,\hat{v}_2)$. Then APLS compares the shortest distance between $(v_1,v_2)$ and $(\hat{v}_1,\hat{v}_2)$. Smaller distance difference means better graph similarity. For both metrics, larger scores indicate better performance.

\subsection{Implementation Details}
In our experiments, all ROIs are $128\times128$-sized. We run the sampling algorithm to collect the training data, and finally obtain around 400K samples to train RNGDet++. During training, the loss weights $\alpha$, $\beta$ and $\gamma$ are set to $5$, $1$ and $1$, respectively. Considering the road topology, we set the number of vertex queries as 10, which is sufficient to handle common road networks. RNGDet++ is trained for 50 epochs, with $10^{-4}$ initial learning rate. All the experiments are conducted on 4 RTX-3090 GPUs.

In our experiments, all approaches are tuned on the validation set of the corresponding dataset. We aim to maximize the TOPO-F1 score of the validation set by trying different parameter settings of the evaluated approach and then use the tuned network to infer the test set. APLS is not considered during parameter tuning.

\subsection{Comparative Experiments}
Our RNGDet++ is compared with seven baseline approaches, including four segmentation-based approaches and three graph-based baselines. The quantitative results are shown in Tab. \ref{tab_comparative_exp}. Qualitative demonstrations on the city-scale dataset are visualized in Fig. \ref{vis}.

Segmentation-based approaches first predict the pixel-level semantic segmentation map of road networks, and then conduct post-processing algorithms to extract and refine the graph of road networks. From Tab. \ref{tab_comparative_exp}, we can see that they tend to have relatively good TOPO scores since TOPO mainly measures the performance of the sub-graph detection, which focuses on the locality. Since segmentation-based approaches directly optimize pixels, they can achieve satisfactory results on pixel-level or local topology-level metrics. However, they have a degraded APLS score mainly because their global topology is not good enough, which could be caused by incorrect detection of road intersections or overlapped overpasses. Therefore, we claim that segmentation-based approaches can detect the road network graph with satisfactory local performance, but cannot present good results on the global scale.

Different from the aforementioned segmentation-based approaches, graph-based approaches directly output and optimize the graph of road networks. Thus, they usually have better topology-level performance. Sat2Graph and RNGDet present the SOTA performance among the baselines, and present superior results not only in the TOPO metrics but also in the APLS metric. RNGDet++ is enhanced from RNGDet by utilizing multi-scale deep features, and it has the best evaluation scores in terms of both APLS and TOPO metrics. So, the superiority and effectiveness of our RNGDet++ are demonstrated and verified.

\begin{table}[t] 
\setlength{\abovecaptionskip}{0pt} 
\setlength{\belowcaptionskip}{0pt} 
\renewcommand\arraystretch{1.0} 
\renewcommand\tabcolsep{1.3pt} 
\centering 
\begin{threeparttable}
\caption{The quantitative comparison results. The best results are highlighted in bold font. 
For all the metrics, larger values indicate better performance.} 
\begin{tabular}{@{} c c c c c c c c c @{}}
\toprule
\multirow{3}{*}{Methods}&\multicolumn{4}{c}{City-scale Dataset} & \multicolumn{4}{c}{SpaceNet Dataset} \\ \cmidrule(l){2-5} \cmidrule(l){6-9}
&Prec. $\uparrow$ & Rec. $\uparrow$ & F1$\uparrow$ & APLS $\uparrow$&Prec. $\uparrow$ & Rec. $\uparrow$ & F1$\uparrow$ & APLS $\uparrow$ \\ 
\midrule
 Seg-UNet \cite{ronneberger2015u} & 75.34 & 65.99 & 70.36 & 52.50 & 68.96& 66.32 & 67.61 & 53.77 \\
 Seg-DRM \cite{mattyus2017deeproadmapper} & 76.54 & 71.25 & 73.80 & 54.32 & 82.79 & 72.56 & 77.34 & 62.26 \\
 Seg-Improved \cite{batra2019improved} & 75.83 & 68.90 & 72.20 & 55.34 &81.56 & 71.38 & 76.13 & 58.82\\
  Seg-DLA \cite{yu2018deep} & 75.59 & 72.26 & 73.89 & 57.22 & 78.99& 69.80& 74.11 &56.36\\
   RoadTracer \cite{bastani2018roadtracer} & 78.00 & 57.44 & 66.16 & 57.29 & 78.61 & 62.45 & 69.90 & 56.03\\
   Sat2Graph \cite{he2020sat2graph} & 80.70 & 72.28 & 76.26 & 63.14 & 85.93&\textbf{76.55}&80.97&64.43\\
    RNGDet \cite{xu2022rngdet} & \textbf{85.97} & 69.78 & 76.87 & 65.75 &90.91 &73.25
&81.13 &65.61\\
    \midrule
   RNGDet++  & 85.65 & \textbf{72.58} & \textbf{78.44} & \textbf{67.76} &\textbf{91.34}&75.24&\textbf{82.51}&\textbf{67.73}\\ 
\bottomrule 
\label{tab_comparative_exp}
\end{tabular} 
\end{threeparttable}
\end{table}

\subsection{Ablation Studies}
We conduct ablation studies to verify the rationality of the design of RNGDet++, including the instance segmentation head and the multi-layer features. The quantitative results of ablation studies are shown in Tab. \ref{tab_ablation}.
 
First, the instance segmentation head is removed from RNGDet++. The instance segmentation head is used to predict road segments ahead of $v_t$, which can better supervise the training of RNGDet++, making it better capture the spatial and topology information of the road network graph. Based on the evaluation scores, RNGDet++ without the instance segmentation head presents inferior results. So, the importance of the instance segmentation head is well demonstrated.

Then, to learn how multi-layer features affect the performance of RNGDet++, we train RNGDet++ by only utilizing the deepest feature layer (i.e., $f_4$). 
From the evaluation results, we find that RNGDet++ without utilizing multi-layer features cannot reach as good results as the original RNGDet++. Thus, the necessity of using multi-scale features is verified.

\begin{table}[t] 
\setlength{\abovecaptionskip}{0pt} 
\setlength{\belowcaptionskip}{0pt} 
\renewcommand\arraystretch{1.0} 
\renewcommand\tabcolsep{3.3pt} 
\centering 
\begin{threeparttable}
\caption{The quantitative results for the ablation study. The best results are highlighted in bold font. For all the metrics, larger values indicate better performance. We assess the instance segmentation head (I) and the multi-scale features (M).} 
\begin{tabular}{c c c c c c c c c c c}
\toprule
 \multirow{3}{*}{I}& \multirow{3}{*}{M}   &\multicolumn{4}{c}{City-scale Dataset} & \multicolumn{4}{c}{{SpaceNet Dataset}} \\ \cmidrule(l){3-6} \cmidrule(l){7-10}
&&Prec. $\uparrow$ & Rec. $\uparrow$ & F1$\uparrow$ & APLS $\uparrow$&{Prec. $\uparrow$} & {Rec. $\uparrow$} & {F1$\uparrow$} & {APLS $\uparrow$}  \\ 
\midrule
 &\checkmark  &\textbf{86.04}&70.65&77.94&66.36&{90.74}&{75.10}&{82.18}&{67.21}\\
 \checkmark&&85.62  &71.88&78.01&66.94&{\textbf{91.46}}&{75.11}&{82.48}&{67.01}\\
 \checkmark&\checkmark & 85.65 & \textbf{72.58} & \textbf{78.44} & \textbf{67.76}&{91.34}&{\textbf{75.24}}&{\textbf{82.51}}&{\textbf{67.73}}\\
\bottomrule 
\label{tab_ablation}
\end{tabular} 
\end{threeparttable}
\end{table}

\subsection{Limitations}

\subsubsection{Lower Efficiency}
Since at each layer, the transformer makes one prediction based on fused features, the time used for the inference of RNGDet++ is relatively longer than RNGDet. The inference time cost is reported in Tab. \ref{tab_time}. However, it should be noted that our task (i.e., road network graph detection) is an offline task, which is not sensitive to efficiency. Thus, considering the superior effectiveness performance of RNGDet++, we think the relatively lower efficiency is acceptable at this stage. We plan to further simplify the network and replace the network modules with lighter structures.

\subsubsection{Too complicated Intersection and Overpass}
Even though RNGDet++ outperforms previous works and presents the best results, it still cannot handle too complicated road intersections or overpasses very well. Moreover, since RNGDet++ is trained by imitation learning, incorrect predictions in these scenarios may affect the subsequent behaviors of the agent. We plan to further optimize the training strategy, and use more powerful backbones to improve the reasoning ability of RNGDet++.

\begin{table}[t] 
\setlength{\abovecaptionskip}{0pt} 
\setlength{\belowcaptionskip}{0pt} 
\renewcommand\arraystretch{1.0} 
\renewcommand\tabcolsep{14pt} 
\centering 
\begin{threeparttable}
\caption{The inference time cost. We report the time cost (hours) used to infer all testing images.} 
\begin{tabular}{@{} c c c c c c c c c c c c c c c@{}}
\toprule
& Sat2Graph&RNGDet&RNGDet++ \\ 
\midrule
City-scale Dataset &2.51h&2.68h&3.85h\\
{SpaceNet Dataset} &{1.15h}&{1.22h}&{1.88h}\\
\bottomrule 
\label{tab_time}
\end{tabular} 
\end{threeparttable}
\end{table}

\section{Conclusions and Future Work}
In this letter, we enhanced the previous road network graph detection approach RNGDet by adding an instance segmentation head, and enabling it to leverage multi-scale features of the backbone. The new novel approach is named RNGDet++. The instance segmentation head could better supervise the network training and improve the reasoning ability of RNGDet++. Besides, RNGDet++ could utilize all layers of features obtained by the backbone, which enables the network to capture multi-scale information of the road network, so that RNGDet++ presents superior results. RNGDet++ is trained by the behavior-cloning imitation learning algorithm. RNGDet++ and all the baselines are fairly evaluated on two large publicly available datasets. Compared with all baselines, RNGDet++ achieves better evaluation scores, not only at the pixel level but also at the topology level. In the future, we plan to further simplify the network structure for efficiency, and apply more powerful backbones to better capture the features of road networks. 

\bibliographystyle{IEEEtran}
\bibliography{mybib}
\end{document}